\documentclass[conference]{article} 
\usepackage{aaai21}  
\usepackage{times}  
\usepackage{helvet} 
\usepackage{courier}  
\usepackage[hyphens]{url}  
\usepackage{graphicx} 
\usepackage{natbib}  
\usepackage{caption} 
\setcitestyle{numbers}
\title{Measuring Difficulty of Novelty Reaction}
\author{
    Ekaterina Nikonova,
    Cheng Xue,
    Vimukthini Pinto,
    Chathura Gamage,
    Peng Zhang,
    Jochen Renz
    \\
}
\affiliations{
    School of Computing, The Australian National University, Canberra, Australia\\

 ekaterina.nikonova, cheng.xue, vimukthini.inguruwattage, chathura.gamage, p.zhang, jochen.renz@anu.edu.au

}

\begin{document}

\maketitle

\begin{abstract}
Current AI systems are designed to solve close-world problems with the assumption that the underlying world is remaining more or less the same. However, when dealing with real-world problems such assumptions can be invalid as sudden and unexpected changes can occur. To effectively deploy AI-powered systems in the real world, AI systems should be able to deal with open-world novelty quickly. Inevitably, dealing with open-world novelty raises an important question of novelty difficulty. Knowing whether one novelty is harder to deal with than another, can help researchers to train their systems systematically. In addition, it can also serve as a measurement of the performance of novelty robust AI systems. In this paper, we propose to define the novelty reaction difficulty as a relative difficulty of performing the known task after the introduction of the novelty. We propose a universal method that can be applied to approximate the difficulty. We present the approximations of the difficulty using our method and show how it aligns with the results of the evaluation of AI agents designed to deal with novelty. 
\end{abstract}

\section{Introduction}
In recent years, significant progress was made in the research of Artificial intelligence (AI) systems for closed-world problems. As the result of this research, AI agents have out-performed humans in many different video games and tasks \cite{Vinyals2019GrandmasterLI, Silver2016MasteringTG}. However, when deployed in the real-world, such AI systems can usually only deal with tasks that closely relate to their previous experience and have poor out-of-domain generalization abilities \cite{Paduraru2021ChallengesOR}. Moreover, if anything changes in underlying task conditions, such systems would need millions of training iterations to adapt to the novel situation. Retraining the system is often costly and requires a lot of data or could be even impossible if an AI system is deployed in the real world. Consider for example a self-driving car that suddenly encounters spilled motor oil on the road. The system would only be allowed to use a very small number of iterations and would have to quickly adapt to the changed road conditions in order to keep its passengers safe. Naturally, pre-training a system on the oily road data would help, however, the number of possible novel situations that the system can encounter is extremely large. Training a system that could deal with all possible scenarios would require an enormous amount of data and computational power. Instead, researchers should focus on designing and creating AI systems that can quickly adapt to novel situations. 

As researchers try to design systems that could efficiently adapt to novel scenarios, it is also important to evaluate such systems. Consider, the example of self-driving cars described above. Assume that we now have two additional systems, one can also deal with oily roads and one that can drive in snowy conditions. Which system is better? Is dealing with sudden oil on the road a harder task than dealing with sudden snow? Can the knowledge of dealing with one novelty be used in dealing with another novelty? When training an AI system which novelty would be the easiest to train on next? All of those questions require us to have a method to measure or approximate the difficulty of the novelty before the agent interacts with it. 

In this paper, we propose a method that can be used to approximate the difficulty of the novelty and predict how hard dealing with this novelty will be for an AI system. We designed different novelties, predicted their difficulty and then evaluated several state-of-the-art novelty agents on these novelties. We compare our difficulty predictions to the actual results of these agents.

\section{Related Work}
\begin{table*}[t]
    \centering
    \begin{tabular}{|c|c|c|c|c|c|c|c|c|c|c|c|c|}
        \hline
        Novelty: & 1 & 2 & 3 & 4 & 5 & 6 & 7 & 8 & 9 & 10 & 11 & 12 \\
        \hline
        Distance: & \textbf{0.76} &\textbf{ 0.46} & \textbf{-0.15} & \textbf{-0.16} & \textbf{0.26} & \textbf{0.29} & \textbf{0.22} & -0.10 & \textbf{0.28} & \textbf{0.43} & 0.23 & -0.03  \\
        Difficulty: & H & H & E & E & M & M & M & E & M & H & M & E \\
        Pass \% Diff.: & \textbf{-35} & \textbf{-13} & \textbf{+20} & \textbf{-5} & \textbf{-7} & \textbf{-9} & \textbf{-19} & -13 & \textbf{-17} & \textbf{-10} & -2 & -7 \\
        \hline
    \end{tabular}
    \caption{Average distance between solution spaces of the sampled pre-novel and post-novel levels, predicted difficulty E-easy, M-medium and H-hard, and average passing rate change (in percentage) of the tested novelty dealing agents. Numbers in bold indicate that predicted and observed difficulty have matched.}
    \label{tab:results}
\end{table*}
Historically, the difficulty of a task is something that is usually defined by some hand-crafted rules \cite{Zhang2021PersonalizedTD} or is agent-dependent. However, when those difficulty rules are unknown or one would like to evaluate several different agents, a method to predict the general agent-independent difficulty of the task is needed. Thus, for example, knowing the difficulty of the task before the agent's interaction can help to select the next task to learn in hierarchical learning \cite{Pateria2021HierarchicalRL}, reinforcement learning \cite{Zhang2021PersonalizedTD} or open-world learning \cite{Bendale2015TowardsOW}. In open-world learning, there were several attempts to formalize the novelty and its difficulty \cite{Boult2020AUF, Muhammad2021ANA}. However, while having good theoretical results, previous approaches have practical limitations such as the need of having an optimal agent \cite{Boult2020AUF} which could be hard to obtain. In this paper, we propose a more practical approach to measure the difficulty. We measure the reaction difficulty in terms of the difference of the solution spaces of a task in previously known and in novel conditions. While measuring the difficulty of a task in terms of the solution spaces has been used before \cite{Hermes2009SolutionSpaceBasedCA, WeissgerberKurt2011DeterminingSS} to the best of our knowledge this is the first work to apply it to measure the difficulty of open-world novelty.

\section{Novelty Reaction Difficulty}
\begin{figure}[h]
\centering
\includegraphics[scale=0.5]{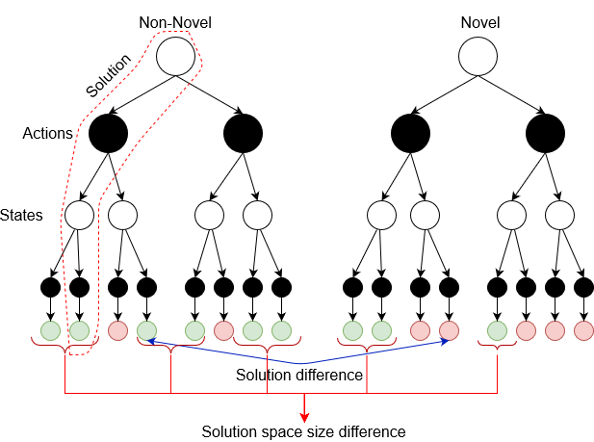}
\caption{An example of solution space difference. Here white nodes are states, black nodes are actions, and the green leaf nodes are final states of a solution.}
\label{fig:sol_space}
\end{figure}

In this paper, we define the reaction difficulty to be a relative difficulty of completing a known task after introducing novelty. We measure the difficulty of the introduced novelty using the distance between pre-novel and post-novel solution spaces (Figure \ref{fig:sol_space}). 

We define $A$ to be a finite pre-novel world that an agent has trained on, and let $B$ be a finite (post-novel) world with the introduced novelty. Let $S$ be a set of all states of the given world, $r(s_n)$ be a binary reward function which assigns $0$ or $1$ to each terminal state $s_n$, and $t=(a_1, a_2,...,a_{n-1})$ be a sequence of actions. 
We define a sequence of actions $t$ to be a \textit{solution} $p$ if and only if that sequence of actions leads to a terminal state $s_n$ such that $r(s_n)=1$.

We define the solution space of the pre-novel world $A$ to be $P_A=\{p^A_1, p^A_2,...,p^A_i\}$ and the solution space of the post-novel world $B$ to be $P_B=\{p^B_1, p^B_2,...,p^B_j\}$, where $i$ may equal to $j$ or not. 

Further, we define a distance between two solutions $p_x^A$ and $p_y^B$ that contains $n$ actions  as:
\begin{equation}
    d_n(p^A_x, p^B_y) = \frac{\sum_{i=1}^n ||a^A_i - a^B_i||}{n * max_a},
\end{equation}

where $max_a$ is the maximum possible distance between two actions, so that $d(p^A_x, p^B_y) \in [0, 1]$. 

In order to compute the difficulty, we first partition the solutions according to the length of action sequences. In particular, we define $P_A = \{P_{A_1},P_{A_2},...P_{A_i} \}$ and  $P_B = \{P_{B_1},P_{B_2},...P_{B_j} \}$, where $P_{A_1}$ means all solutions of the pre-novel world $A$ with only one action. 

We then define the distance between each partition as:
\begin{equation}
    D(P_{A_n}, P_{B_n}) = \frac{(|P_{A_n}| - |P_{B_n}|) + \sum_{i,j} d_n(p^{A_i}, p^{B_j})}{max(|P_A|, |P_B|)},
\end{equation}
where $i \in \{1,2,...,min(|P_A|, |P_B|)\}$, and where $j \in \{ j| ~ d_n(p^{A_i}, p^{B_j}) = min(d_n(p^{A_i}, p^{B_n})) \forall p^{B_n} \in P_{B_n} \}$.

Finally, given all of the definitions above, we define the {\em novelty reaction difficulty} as the distance between two solution spaces $P_A$ and $P_B$:
\begin{equation}
    D(P_A, P_B) = \frac{\sum_{n}  D(P_{A_n}, P_{B_n})}{\max(|P_A|, |P_B|)}
\end{equation}
which measures the distance between the solution spaces in terms of the solution space shifts and size differences. Our distance measure was inspired by Hausdorff distance. However, in our case the set size difference is also important as if the novelty reduces the number of possible solutions it should be considered to be more difficult than the novelty that does not. As Hausdorff distance does not consider this difference, we propose our own distance function that is similar to Hausdorff distance but also considers the size difference.

Given a finite world, we can approximate the solution space using random sampling agents. As the number of sampling agents increases (i.e. $n \rightarrow \infty$), the entire solution space will be explored. However, often, the solution space is too large to explore entirely and only approximation is possible. In this paper, we use 50 randomly initialized agents that explore the pre-novel and post-novel solution spaces for 100 episodes. We then use the collected solutions as approximations of the solution spaces of pre-novel and post-novel worlds and compute the distance between them.

\section{Experiments}

In our experiments, we use Science Birds as the testing environment. Science Birds is a clone of Angry Birds - a popular video game. The goal of  Science Birds is to destroy all pigs in the game level while also maximizing the overall score. In our setting, we have a number of generated Science Birds levels. We then randomly sample those levels to produce a set of "base" levels. We then introduce a predefined novelty to those base levels and use them to estimate the difficulty. 

In our evaluation, we have a total of 12 novelties that we test. To measure the difficulty of each novelty we have randomly sampled 100 base levels and introduce the novelty to each level. Then given two sets of levels with and without novelty, we run $n$ random agents on each pair of the levels to collect the solutions for the post and pre-novel levels. Once the solutions are collected, we compute the distance between the solution spaces using our method and classify 12 novelties into easy, medium, and hard difficulty. We define a novelty to be easy when $D(P_A, P_B) <= 0$, medium when $0 < D(P_A, P_B) <= 0.3$ and hard otherwise.

Once the novelties are assigned their difficulty, we run state-of-the-art novelty reaction agents on all novelties using a bigger set of levels ($>1000$ levels per novelty) to see if their performance matches our difficulty predictions.

The results for each of the 12 evaluated novelties are displayed in Table \ref{tab:results}. As it can be seen in Table \ref{tab:results} the average distance between the solution spaces and the predicted difficulty matches the observed passing rates of the agents in the majority of the cases. We note that while two novelties, novelty 8 and novelty 11 were classified as easy and medium respectively, the actual agent's evaluation showed that their difficulties should be swapped. We hypothesize that those two novelties were misclassified due to the randomness of the sampling agents or due to the limited number of levels that were sampled to measure the difficulty. We hypothesize that increasing the number of sampling agents or the number of sampled levels will increase the accuracy of the approximations and difficulty predictions. Another possible improvement would be to use a more sophisticated exploration strategy to sample the solution spaces.

\begin{table}[t]
    \centering
    \begin{tabular}{|c|c|c|c|}
        \hline
        Difficulty: & Easy & Medium & Hard  \\
        \hline
        Pass \% in novel games: & 0.3946 & 0.3874 & 0.2214 \\
        \hline
    \end{tabular}
    \caption{Average pass percentage in novel game against predicted difficulty of the novelty.}
    \label{tab:resultsavg}
\end{table}

Despite misclassification of individual novelties difficulties, on average, the predicted level of difficulty matched the average passing rate as shown in Table \ref{tab:resultsavg}. 

Thus, we can conclude that despite a relatively small number of sampled levels, our method was able to predict the difficulty of the novelties for the agents with high accuracy.

\section{Conclusions}
In this paper, we have proposed a solution space-based method to measure the difficulty of the novelty. We defined novelty reaction difficulty as a relative difficulty of dealing with the previously known task after the introduction of the novelty. We have evaluated our difficulty approximation method on several novelties and compared the predicted difficulties to the passing rates of the agents that are able to adapt to the novelty. The empirical results show that on average there is a correlation between agent's passing rates and the predicted difficulty. Moreover, our results show that the average passing rates decrease as the predicted difficulty increases. 

While solution space-based analysis was used in previous work, to the best of our knowledge were are the first to apply it to determine the difficulty of open-world novelties. Compared to the previous approach, our difficulty measure is more practical and can be easily applied to other domains where the difficulty of the novelty can be measured by the difference in the solution spaces.

\end{document}